\documentclass{article}
\usepackage{spconf,amsmath,graphicx}
\usepackage{booktabs}
\usepackage{multirow}
\usepackage{tabularx}
\usepackage{enumitem}

\title{DYNAMIC WEIGHT-BASED TEMPORAL AGGREGATION FOR LOW-LIGHT VIDEO ENHANCEMENT UNDER EXTREME NOISE}
\name{Ruirui Lin, Guoxi Huang, Nantheera Anantrasirichai\thanks{This work was supported by UKRI MyWorld Strength in Places Programme (SIPF00006/1).}}
\address{Visual Information Laboratory, University of Bristol, United Kingdom}
\begin{document}
%
\maketitle
\begin{abstract}
Low-light video enhancement (LLVE) is challenging due to noise, low contrast, and color degradation. While learning-based methods enable fast inference, they often fail under heavy real-world noise because they do not sufficiently exploit long-term temporal cues. We propose DWTA-Net, a novel deep-learning recurrent LLVE framework with a recurrent design. DWTA-Net adopts an integrated two-stage architecture: Stage I restores local structure and color via multi-frame alignment for temporally consistent Mamba-based enhancement, while Stage II performs recurrent refinement using a novel dynamic weight-based temporal aggregation guided by optical flow, functioning as a recurrent denoiser that adapts to motion. We further introduce a texture-adaptive loss that preserves fine details in textured regions while suppressing noise in homogeneous areas. Experiments on real-world low-light footage show that DWTA-Net achieves stronger noise suppression and fewer artifacts, delivering superior visual quality compared with state-of-the-art methods.
\end{abstract}
\section{Introduction}
\label{sec:intro}

Capturing high-quality video in low-light environments remains a fundamental challenge in computer vision. Applications ranging from autonomous surveillance to consumer photography suffer when the signal-to-noise ratio (SNR) drops, leading to severe degradations such as loss of contrast, color shift, and strong sensor noise~\cite{zheng2024lowlightsruvey}. These challenges are amplified in outdoor scenes in the wild, where uneven illumination, motion, and complex sensor noise further complicate restoration, making traditional pipelines inadequate.

While significant progress has been made in single-image low-light enhancement, these solutions often fail when applied to video sequences. The primary challenge is the temporal inconsistency, as processing frames in isolation inevitably leads to inter-frame flickering and ghosting artifacts. To mitigate this, recent \textbf{Low-Light Video Enhancement (LLVE)} research has shifted toward multi-frame modeling~\cite{Lin:STA:2024,Jiang:learn:2019}. However, many current architectures rely on sliding-window processing or 3D convolutions that only consider a short neighborhood of frames. Such methods are often computationally heavy, require a large memory, and, more importantly, cannot fully leverage long-term temporal redundancy, which is crucial for denoising under extreme low-light noise.

\begin{figure}[t!]
\centering
\vspace{-0.1mm}
\includegraphics[width=\columnwidth]{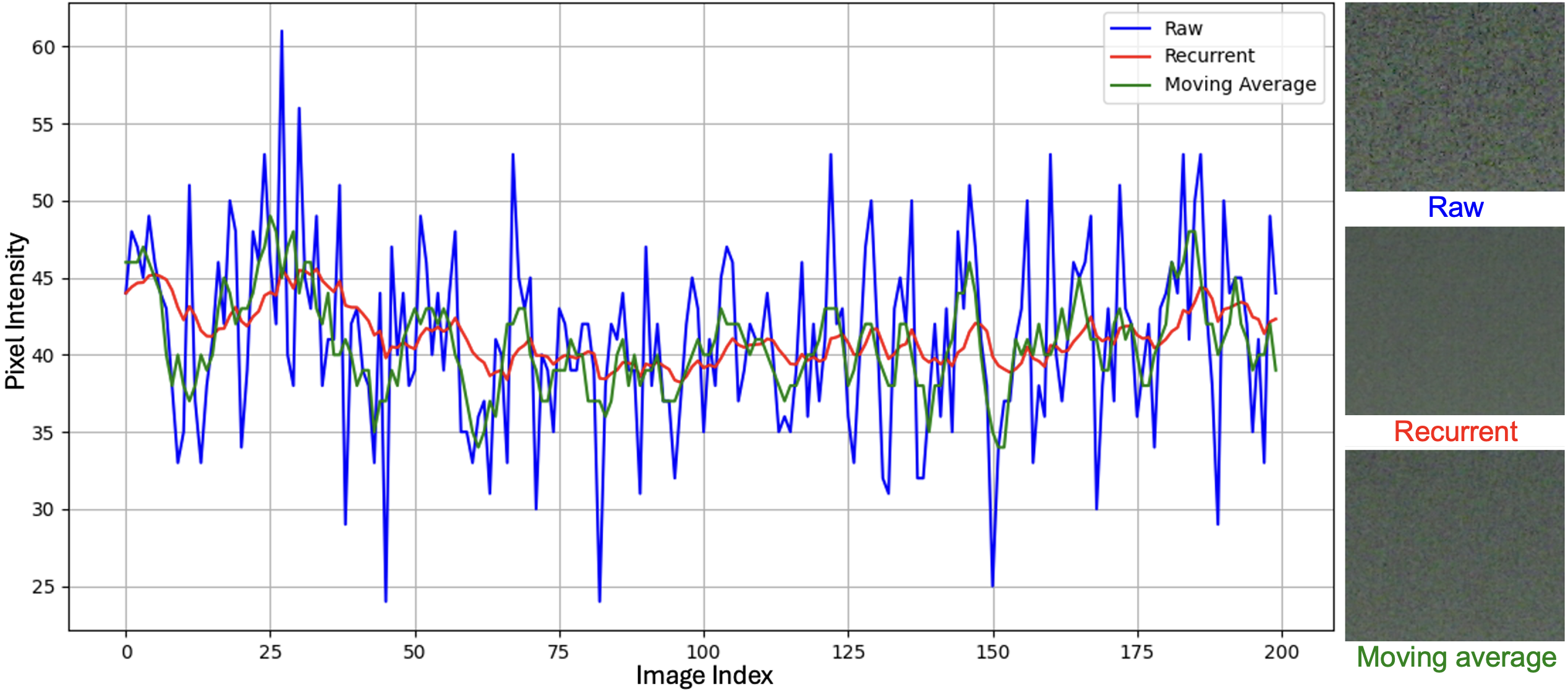}
\caption{Noise suppression comparison: recurrent aggregation vs. fixed 5-frame averaging. Recurrence leverages long-term information for stronger suppression.}
\label{fig:noise_reduction_method}
\vspace{-3mm}
\end{figure}

Beyond these temporal modeling limits, robustness is also constrained by the noise distributions seen during training. Real-world low-light noise is typically more complex and severe than training noise, so models often generalize poorly to in-the-wild footage, especially when extreme noise dominates outdoor scenes. This motivates a design that can adaptively suppress spatially varying noise while preserving fine structures over long sequences.

This paper proposes the \textbf{D}ynamic \textbf{W}eight-based \textbf{T}emporal \textbf{A}ggregation \textbf{N}etwork (DWTA-Net), a two-stage Mamba-based framework tailored to heavy real-world noise. Stage I utilizes Mamba, providing global context that stabilizes brightness, color, and coarse structure in severely degraded sequences. Stage~II departs from conventional parallel-frame fusion by introducing a recurrent formulation that accumulates long-term temporal evidence. Motivated by the observation that averaging across time is a natural denoiser~\cite{hassan2010mechanical,lehtinen2018noise2noise}; accordingly, DWTA-Net recursively aggregates past outputs to directly suppress noise while promoting temporal consistency via optical-flow alignment. We employ the recurrent mechanism in the spatial space rather than the feature space, since noise predominantly manifests as high-frequency artifacts in low-level vision. To balance noise suppression and motion sharpness, we design the recurrent update with an exponential decay weighting strategy and motion-residual guidance, enabling stronger smoothing in static regions while avoiding blur in dynamic content. As shown in Fig.~\ref{fig:noise_reduction_method}, the recurrent formulation produces significantly cleaner results than fixed-window averaging, supporting the need for long-term temporal aggregation under heavy noise. Finally, we introduce a texture-adaptive loss that preserves fine details in textured areas while promoting smoothness in homogeneous regions.

Our main contributions are summarized as follows:
\begin{itemize}[noitemsep, topsep=0pt, leftmargin=*]
    \item We propose a novel deep-learning, Mamba-based, recurrent framework for LLVE, DWTA-Net, featuring a recurrent design tailored to heavy real-world noise.
    \item We develop a spatially adaptive, dynamic weight-based temporal aggregation strategy. By utilizing motion residuals, the model adaptively adjusts blending weights to balance temporal consistency with motion-aware sharpness.
    \item we further design a texture-adaptive loss that balances fine detail preservation with spatial smoothness.  
\end{itemize}           

 \begin{figure*}[ht!]
 \vspace{-0.1mm}
    \centering
    \includegraphics[width=\linewidth]{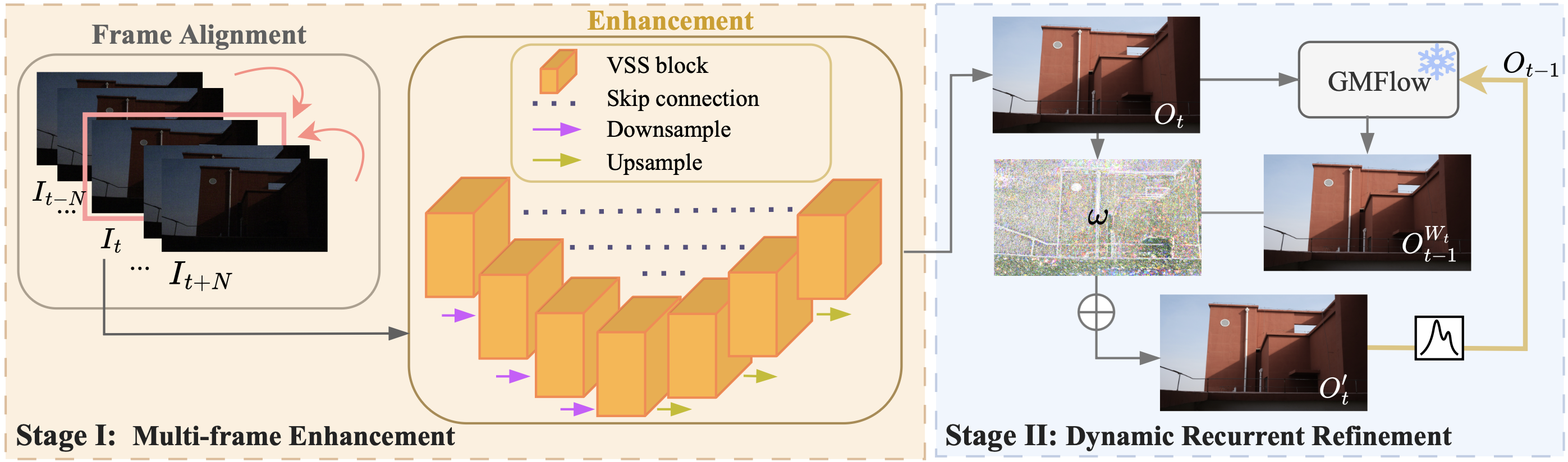}
    \caption{Overview of the proposed DWTA-Net. (a) Stage I: multi-frame enhancement for brightness and structure restoration. (b) Stage II: recurrent refinement with dynamic temporal aggregation for long-term consistency and heavy-noise denoising.}
    \label{fig:full_arch}
\vspace{-3mm}
\end{figure*}

\section{Related Work}

\subsection{Low-Light Image Enhancement (LLIE)}
The evolution of LLIE can be categorized into traditional and learning-based eras. Traditional methods relied mainly on Retinex theory~\cite{Land1977TheRT}, which decomposes an image into reflectance and illumination. Approaches such as Histogram Equalization~\cite{ibrahim2007brightness} and BM3D~\cite{bm3d} provided early benchmarks, but often struggled with over-amplification and manual parameter tuning. 

With deep learning, CNNs enabled data-driven restoration. Recent work has shifted to higher-capacity models: Transformers~\cite{Cai_2023_ICCV} for long-range spatial modeling, Mamba-based State-Space models~\cite{zou2024wavemamba} for wide receptive fields with efficiency, and diffusion models~\cite{jiang2023low} for realistic texture generation. However, these methods still lack mechanisms to enforce temporal consistency in videos.

\subsection{Low-Light Video Enhancement (LLVE)}
To address the temporal dimension, LLVE methods must align and aggregate information across frames. Early approaches used 3D convolutions to learn spatiotemporal features directly~\cite{Jiang:learn:2019}. Later, hybrid frameworks combined Retinex decomposition with self-supervised denoising modules~\cite{wang2021sdsd}, while multi-input denoisers were developed for extremely low-light scenarios under starlight~\cite{Monakhova_2022_CVPR}.

A significant trend in LLVE is the use of explicit motion compensation. Methods such as~\cite{Lin:STA:2024, Lin:LL:2024} utilize deformable convolutions (DCN) to align neighboring frames before fusion. While effective for short-term motion, these sliding window methods are limited by the number of input frames they can process simultaneously due to memory constraints. Our DWTA-Net differs by using a recurrent state, allowing it to remember information from a number of past frames without a proportional increase in computational cost. 

\subsection{Recurrent Video Restoration}
Recurrent models and their variants have been successful in tasks like Super-Resolution and general denoising~\cite{Liang_NEURIPS2022}. These models propagate hidden states to maintain long-term consistency. In low-light restoration, recurrent designs have been explored for single-image iterative refinement~\cite{Yang_2020_CVPR} and binarized raw video enhancement~\cite{BRVE_2024_CVPR}.

However, applying recurrence to low-light \textit{video} introduces a unique challenge: the propagation of noise. In low-SNR regimes, a naive recurrent model may accumulate errors or create over-smoothing artifacts. Unlike latent recurrence, our method directly aggregates in pixel/spatial space, where low-light noise mainly appears as high-frequency artifacts; motion residuals further guide adaptive smoothing in static regions while preserving dynamic details.

\section{Methodology}
\label{sec:method}

\subsection{DWTA-Net}
Our DWTA-Net enhances low-light videos in two stages, as shown in Figure \ref{fig:full_arch}:  
(1) multi-frame alignment and enhancement for brightness and structure restoration, and  
(2) recurrent refinement with dynamic temporal aggregation for long-term consistency.  

\subsubsection{Stage I: Multi-frame Enhancement}  
This stage addresses short-term temporal consistency and performs initial restoration of brightness, color, and structure. To reduce flickering, a short sequence of neighboring frames is first passed through the PCD module~\cite{wang2019edvr}, which aligns them to a reference and outputs motion-compensated features. These aligned features are then processed by a Mamba-based U-Net–like backbone where conventional convolutional blocks are replaced with Visual State-Space (VSS) blocks~\cite{liu2024vmamba}. Unlike standard convolutions with limited receptive fields, VSS blocks utilize a selective scan mechanism to capture long-range global dependencies, which is critical for restoring structural coherence in heavily degraded scenes. Formally, given an input feature $\mathbf{h}_{l-1}$, the update at layer $l$ is defined as:
\begin{equation}
    \begin{aligned}
        & \mathbf{h}_l = \mathrm{SS2D}\!\left(\mathrm{LN}(\mathbf{h}_{l-1})\right) + \mathbf{h}_{l-1},\\
        & \mathbf{h}_{l+1} = \mathrm{FFN}\!\left(\mathrm{LN}(\mathbf{h}_l)\right) + \mathbf{h}_l,
    \end{aligned}
\end{equation}
where SS2D is the selective-scan operator, and FFN is a feedforward layer. The Stage~I output frame is denoted as $O_t$.

\subsubsection{Stage II: Dynamic Recurrent Refinement.}  
Stage II aggregates information over time to suppress noise and stabilize details. Departing from hidden-state recurrence, we perform aggregation directly in the spatial space to specifically target high-frequency noise artifacts. This stage primarily relies on a motion-based, dynamic weighted blending process. At each timestep $t$, the refined output from the previous step $O'_{t-1}$ is aligned to the current Stage~I output $O_t$ using optical flow  (GMFlow~\cite{xu2022gmflow} is employed in this paper), yielding $O^{W_t}_{t-1}$ as shown in Figure \ref{fig:full_arch}. At initialization ($t=0$), we set $O'_0 = O_0$. To improve flow estimation under varying illumination, we apply a brightness adjustment to $O'_{t-1}$ using $O_t$ as a reference. 

To adaptively balance static and dynamic regions, we compute a dynamic weight map $\omega$ based on the residual $R = |O_t - O^{W_t}_{t-1}|$. This dynamically controls the influence of previous frames. $\omega$ is calculated using a weighted sigmoid smoothing function:
\begin{equation}
    \omega = c + \frac{1-c}{1+\exp(-a(R-b))},
\end{equation}
where $a$ controls steepness, $b$ sets the residual threshold, and $c$ defines the minimum contribution of the warped frame. In our experiments, we set $(a, b, c) = (10, 0.5, 0.1)$, chosen empirically based on validation performance. In static regions (low $R$), $\omega \approx c$ emphasizes accumulation from $O^{W_t}_{t-1}$, while in dynamic regions (high $R$), $\omega \approx 1$ favors the current frame $O_t$. The refined output is
\begin{equation}
    O'_t = \omega \cdot O_t + (1-\omega) \cdot O^{W_t}_{t-1}.
\end{equation}
This recurrent update is propagated to the next step, enabling long-range temporal integration. The dynamic weighting ensures stability in static areas while preserving details in motion regions, overcoming the limitations of fixed-window averaging or latent-space recurrence.

\begin{figure*}[t!]
    \centering
    \small
    \includegraphics[width=\linewidth]{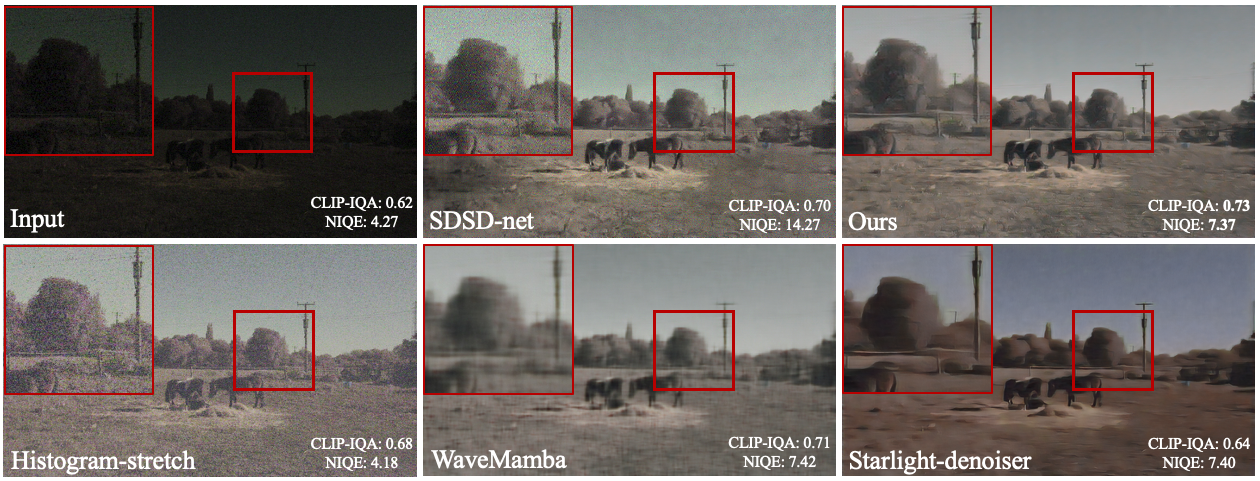}
    \caption{low-light enhancement comparison using histogram stretching, SDSD-net, Starlight, WaveMamba, and our method. }
    \label{fig:vis_enhance_comp}
\end{figure*}

\subsection{Proposed Texture-Adaptive Loss}
Low-light noise is spatially variant. Textured regions require preservation of detail, whereas smooth regions benefit from stronger denoising. We introduce a texture-aware map $M_T \in [0,1]$ derived from high-frequency components of $O'_t$ via a one-level 2D Discrete Wavelet Transform (2D-DWT), extracting horizontal, vertical, and diagonal sub-bands. These sub-bands are then concatenated into a high-frequency representation $O_{high}$. The result is normalized to serve as a measure of texture complexity:
\begin{equation}
M_{T} = \frac{M_\mu^\sigma - \min(M_\mu^\sigma)}{\max(M_\mu^\sigma) - \min(M_\mu^\sigma)}, \: M_\mu^\sigma =\mu\left(\sigma(O_{\text{high}})\right),
\end{equation}

\noindent The proposed texture adaptive loss, $\mathcal{L}_{text-adaptive}$, is defined as follows: 
\begin{equation}
    \mathcal{L}_{text-adaptive} = \big[ M_T \odot \mathcal{L}_{VGG} + (1-M_T) \odot \mathcal{L}_{TV} \big],
\end{equation}
where $\odot$ denotes element-wise multiplication. In highly textured regions (where $M_T \to 1$), we apply perceptual loss $\mathcal{L}_{VGG}$~\cite{VGG} to enforce feature-level similarity and preserve fine details. In homogeneous regions (where $M_T \to 0$), we encourage smoothness and suppress noise using the Total Variation (TV) loss~\cite{TV}, which penalizes sharp gradients. The texture-aware map $M_T$ weights these two components. 

\noindent The overall loss is computed between the final output $O'_t$ and its corresponding ground truth $GT_t$. It consists of a pixel loss and a weighted texture-adaptive term, where we set $\alpha = 0.5$ (chosen on the validation set). The pixel loss $\mathcal{L}_{pixel}$ is an $\ell_2$ reconstruction loss that preserves the overall structure and content:
%
\begin{equation}
    \mathcal{L} = \mathcal{L}_{pixel} + \alpha \cdot \mathcal{L}_{text-adaptive}.
\end{equation}

\section{Experiments}

\subsection{Experimental Settings}

DWTA-Net is trained on the paired low-light video dataset DID~\cite{Fu_DID}. Existing LLVE datasets are often limited in quality and scale; we therefore select DID, as it is a high-quality dataset that provides genuine low-light noise for training.

While we report quantitative results on this dataset using full-reference metrics, our primary goal is to evaluate the model's effectiveness in practical, unconstrained scenarios with heavy noise. To this end, we focus our qualitative evaluation on challenging in-the-wild low-light videos that are entirely separate from the training data. In particular, we highlight the \textit{Horse} sequence, a professional filming dataset captured after sunset with a Canon ML-105. This serves as a comprehensive benchmark due to its diverse degradations, including homogeneous skies, textured grass, and fine structural details such as fences and electric cables. 

The proposed model is trained for $3\times 10^5$ iterations using NVIDIA RTX 3090/5090 GPUs with 5 input frames for the alignment in Stage~I. The Adam optimizer is adopted for optimization with an initial learning rate of $1\times 10^{-4}$. The batch size is set to 1 and the patch size to $512\times512$. 

For the paired dataset (DID) used for training, we use three full-reference metrics: PSNR and SSIM to assess fidelity, and LPIPS to measure perceptual quality. For real-world (unpaired), in-the-wild footage, we adopt the NIQE and CLIP-IQA as no-reference metrics. Higher CLIP-IQA and lower NIQE are better. 

\begin{table}[t]
\centering
\small

\begin{tabular*}{\columnwidth}{@{\extracolsep{\fill}}c|ccc}
\toprule
Method & PSNR $\uparrow$ & SSIM $\uparrow$ & LPIPS $\downarrow$  \\
\midrule
Retinexformer~\cite{Cai_2023_ICCV}$^*$     & \underline{24.15}  & 0.849  & 0.216 \\
DiffLL~\cite{jiang2023low}$^*$            & 21.03 & 0.753 & \underline{0.117} \\ 
WaveMamba~\cite{zou2024wavemamba}$^*$     & 21.31 & 0.748  & 0.513\\
\midrule
EDVR~\cite{wang2019edvr}                  & 22.91 & 0.785 & 0.199\\
Starlight-denoiser~\cite{Monakhova_2022_CVPR} & 19.12 & 0.731 & 0.237\\
SMOID~\cite{Jiang:learn:2019}             & 21.71 & \textbf{0.880} & 0.194 \\
SDSD-net~\cite{wang2021sdsd}              & 21.88 & 0.834 & 0.216 \\
\midrule
DWTA-Net                                  & \textbf{24.27} & \underline{0.857} & \textbf{0.115}\\
\bottomrule
\end{tabular*}

\vspace{1mm}
(A) Enhancement performance comparison on DID dataset

\vspace{3mm}

\begin{tabular*}{\columnwidth}{@{\extracolsep{\fill}}c|ccc}
\toprule
Method & PSNR $\uparrow$ & SSIM $\uparrow$ & LPIPS $\downarrow$\\
\midrule
Retinexformer~\cite{Cai_2023_ICCV}$^*$     & \underline{27.72} & \underline{0.833} & 0.566   \\
DiffLL~\cite{jiang2023low}$^*$            & 27.18 & 0.809 & \textbf{0.170} \\ 
WaveMamba~\cite{zou2024wavemamba}$^*$     & 26.75 & 0.815 & 0.432 \\
\midrule
EDVR~\cite{wang2019edvr}                  & 23.71 & 0.808 & 0.213\\
Starlight-denoiser~\cite{Monakhova_2022_CVPR} & 19.35 & 0.803 & 0.214\\
SMOID~\cite{Jiang:learn:2019}             & 17.04 & 0.703 & 0.340\\
SDSD-net~\cite{wang2021sdsd}              & 23.30 & 0.300 & 0.421 \\
\midrule
DWTA-Net                                  & \textbf{27.81} & \textbf{0.839} & \underline{0.210}\\
\bottomrule
\end{tabular*}

\vspace{1mm}
(B) Denoising performance comparison on \textit{Horse} dataset

\caption{(A) Enhancement tested on paired dataset. (B) Supplementary denoising on the static sky region of the \textit{Horse} footage. $^*$ denotes image-based methods. The best results are highlighted in \textbf{bold} and the second-best results are \underline{underlined}.}
\label{tab:enhance_compare}
\end{table}

\subsection{Performance Comparison}

As video-based methods for lowlight remain limited, we also compared DWTA-Net against state-of-the-art image-based methods. Self-supervised methods typically underperform supervised ones; therefore, we compare only supervised methods for fair comparison. For quantitative comparison, all models were trained and evaluated on the same paired dataset - DID. As shown in Table~\ref{tab:enhance_compare}~(A), DWTA-Net achieves superior fidelity and perceptual quality, attaining the highest PSNR and LPIPS while remaining competitive in SSIM. 

Beyond these quantitative gains, we present additional visual results on a selected in-the-wild video sequence with heavy noise. Since no ground truth is available, we apply histogram stretching to the input to better visualize noise and compare outputs. Fig.~\ref{fig:vis_enhance_comp} shows our model preserving fine detail, texture, and natural color without artifacts, consistent with no-reference metrics (CLIP-IQA and NIQE). Human perception, however, reveals that Starlight-denoiser over-smooths textured regions such as trees and grass, causing detail loss and color distortion, while SDSD-net introduces distracting artifacts in complex grass regions.

\subsection{Denoising performance}
  
\begin{figure}[t]
    \centering
    \includegraphics[width=\linewidth]{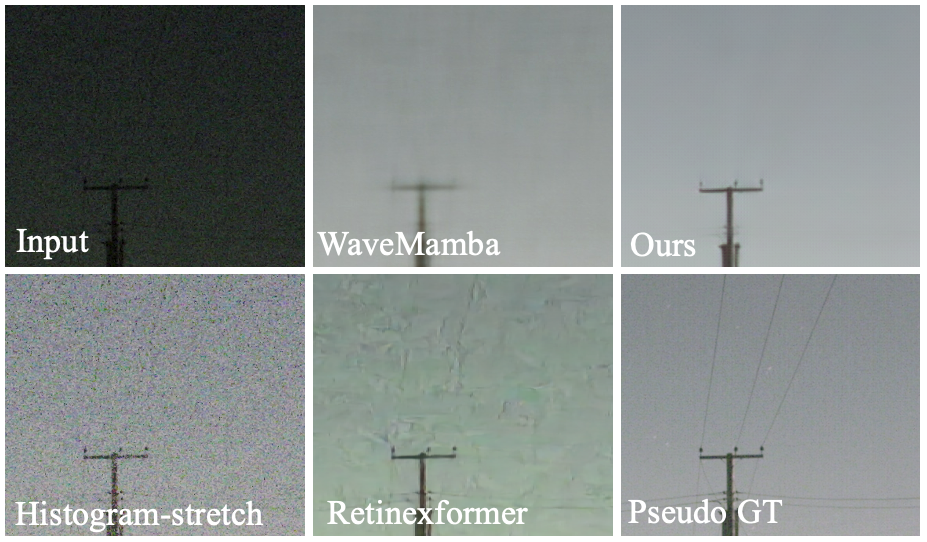}
    \caption{Qualitative comparison of denoising performance (and brightness enhancement) on the static sky. }
    \label{fig:vis_denoise_com}
\end{figure}

Evaluating real-world denoising is challenging as in-the-wild footage lacks ground-truth (GT), and no paired video datasets capture comparable heavy noise. Following prior works such as~\cite{Monakhova_2022_CVPR} that generate pseudo-GTs for noise evaluation, we constructed a supplementary pseudo-paired benchmark by averaging frames from a static region (the sky) of the \textit{Horse} scene. Fig.~\ref{fig:vis_denoise_com} shows that the pseudo-GT is sufficiently clean as a reference. 

Table~\ref{tab:enhance_compare}~(B) compares DWTA-Net against both video and image-based methods; the latter are included for a comprehensive comparison as video-based benchmarks are limited. Remarkably, DWTA-Net outperforms state-of-the-art image-based methods like WaveMamba and Retinexformer in their ideal static scenario, achieving the highest PSNR and SSIM. As illustrated in Fig.~\ref{fig:vis_denoise_com}, none of the compared methods can fully recover extremely fine structures (e.g., electrical cables) under severe noise. Nevertheless, DWTA-Net produces noticeably smoother denoising results with fewer visual distortions than the pseudo-GT reference.

\subsection{Ablation Study}
\begin{table}[t]
\centering
\small

\begin{tabular*}{\columnwidth}{@{\extracolsep{\fill}}c|ccc}
\toprule
  & PSNR $\uparrow$ & SSIM $\uparrow$ & LPIPS $\downarrow$\\
\midrule
w/o Stage~I  &  7.61 & 0.233 & 0.595\\
w/o Stage~II & 23.21 & 0.815 & 0.234\\
\midrule
Proposed & \textbf{24.27} & \textbf{0.857} & \textbf{0.115}\\
\bottomrule
\end{tabular*}

\vspace{1mm}
(A) DWTA-Net

\vspace{3mm}

\begin{tabular*}{\columnwidth}{@{\extracolsep{\fill}}c|ccc}
\toprule
  & PSNR $\uparrow$ & SSIM $\uparrow$ & LPIPS $\downarrow$\\
\midrule
w/o $\mathcal{L}_{texture-adaptive}$ & 23.39 & 0.829 & 0.193\\
w/o $\mathcal{L}_{pixel}$            & 12.56 & 0.704 & 0.343\\
\midrule
Proposed & \textbf{24.27} & \textbf{0.857} & \textbf{0.115}\\
\bottomrule
\end{tabular*}

\vspace{1mm}
(B) Loss function

\caption{Ablation study on the effectiveness of (A) each module of DWTA-Net, and (B) different loss functions.}
\label{tab:abaloss}
\end{table}

\subsubsection{DWTA-Net} We evaluate the effectiveness of each primary module of our network by removing one at a time, as shown in Table~\ref{tab:abaloss}~(A). Removing the Stage~I multi-frame enhancement module leads to poor color and brightness restoration. This loss of basic restoration causes a drastic decrease in PSNR and SSIM, while the high LPIPS indicates severely degraded perceptual quality. Without the Stage~II dynamic recurrent refinement module, the model can only leverage short-term temporal information. While PSNR and SSIM remain acceptable, the increase in LPIPS indicates degraded perceptual quality, evidencing the importance of temporal aggregation for suppressing noise and improving visual quality.

\subsubsection{Loss Function} We further evaluate the impact of each loss component in Table~\ref{tab:abaloss}~(B). Training with only pixel loss (without texture-adaptive loss) yields reasonable results but poorer perceptual quality, as indicated by a higher LPIPS value, suggesting difficulty in balancing noise removal and texture preservation. Whereas using only the proposed texture-adaptive loss (without pixel loss) results in poor overall performance, as the model fails to reconstruct image content accurately without pixel-wise reconstruction loss.


\section{Conclusion}
In summary, DWTA-Net delivers robust low-light video enhancement via a two-stage recurrent framework. Stage~I restores structure and color via short-term multi-frame alignment, while Stage~II refines results through dynamic weight-based temporal aggregation guided by optical flow to exploit long-term temporal evidence. The texture-adaptive loss further improves perceptual quality by balancing detail preservation and smoothness. Extensive benchmarks and challenging in-the-wild evaluations demonstrate that DWTA-Net achieves state-of-the-art performance in suppressing heavy real-world noise, validating its dynamic, motion-aware aggregation as a practical solution for video enhancement.

\vfill\pagebreak

\bibliographystyle{IEEEbib}
\bibliography{main}

\end{document}